\pdfoutput=1

\documentclass[11pt]{article}

\usepackage[preprint]{acl}

\usepackage{times}
\usepackage{latexsym}

\usepackage[T1]{fontenc}

\usepackage[utf8]{inputenc}

\usepackage{microtype}

\usepackage{inconsolata}

\usepackage{graphicx}

\usepackage{enumitem}
\usepackage{arydshln}
\usepackage{graphicx}
\usepackage{amsmath}
\usepackage{color, colortbl,booktabs}
\definecolor{Gray}{gray}{0.9}
\usepackage{linguex}
\newcommand{\target}{\textit{target}}
\newcommand{\cs}{\textit{CS}}
\newcommand{\dataset}{\textsc{SuperSem}}

\title{Superlatives in Context: \\  Modeling the Implicit Semantics of Superlatives}

 \author{ Valentina Pyatkin$\thanks{~~This work was completed in partial fulfillment for the PhD degree of Valentina Pyatkin.}$\textsuperscript{\,\,\,1,2} \,
 Bonnie Webber\textsuperscript{3} \,
 Ido Dagan\textsuperscript{1} \, 
 Reut Tsarfaty\textsuperscript{1}\\\
\textsuperscript{1}Bar Ilan University \\
\textsuperscript{2}Allen Institute for AI \\
\textsuperscript{3}The University of Edinburgh\\
 \texttt{valentinap@allenai.org}
}

\begin{document}
\maketitle
\begin{abstract}
{\em Superlatives} are used to single out elements with a maximal/minimal property. Semantically, superlatives perform a set comparison: \textit{something} (or some things) has the min/max \textit{property} out of a \textit{set}. As such, superlatives provide an ideal phenomenon
for studying implicit phenomena and discourse restrictions.
While this comparison set is often not explicitly defined, its (implicit) restrictions can be inferred from the discourse context the expression appears in. 
In this work we provide an extensive computational study on the semantics of superlatives. We propose a unified account of superlative semantics which allows us to derive a broad-coverage annotation schema. Using this unified schema we annotated a multi-domain dataset of superlatives and their semantic interpretations. We specifically focus on interpreting implicit or ambiguous superlative expressions, by analyzing how the discourse context restricts the set of interpretations. 
In a set of experiments we then analyze how well models perform at variations of predicting superlative semantics, with and without context. We show that the fine-grained semantics of superlatives in context can be challenging for contemporary models, including GPT-4. 
\end{abstract}

\section{Introduction}
Superlatives are used to express a certain type of comparison in language.
They work as domain-based comparisons: an expression like  ``the smallest fish'' means that there is a fish which is smaller than all other fish in a specific set. An interpretation of the superlative comparison requires the human or machine to identify the \target, e.g., the entity or event being the {\em max} or {\em min} of a set, and the \textit{comparison set} (CS), e.g., the set of entities or events against which you are comparing the \target.

Appropriately defining the comparison set requires understanding the general domain, and it is often essential to read beyond the sentence level, or to draw inferences from world knowledge. Take, for example, the following statement:

\ex. \label{firstex} \small{Tom went fishing at the lake together with his friends. He caught the \textbf{largest} fish.}

In ex.~\ref{firstex} the sentence with the superlative `largest' does not provide enough information to properly define the CS, except that one is comparing a fish to other fish.
With the help of the previous sentence, one can restrict the comparison set to the \textit{fish that were caught by Tom and his friends at the lake}.
In this paper we propose that recognizing the CS hinges on identifying the relevant entities or events from context, which can appear both before or after the expression. Specifically here, the  \textit{catching} event is crucial in restricting the CS. 

Being able to automatically interpret the semantics of superlatives can be useful for many downstream applications, such as dialogue state tracking or mining product reviews \cite{scheible2010smallest, bakhshandeh2016learning}. They appear in semantic parsing datasets, like text-to-SQL ('book the earliest flight to Boston' \cite{price1990evaluation}) and their accurate semantic representation might improve Question Answering or Information Extraction. 

To the best of our knowledge, superlatives are understudied in NLP, and so far there has been no systematic work on automatically identifying the CS restrictions from the larger discourse. While \citet{scheible2008annotating, scheible2012textwiki} annotated the comparison set of one semantic superlative subtype, they did so only when it is explicitly expressed in the syntactic construction of the sentence. Similarly, \citet{bos2006empirical} mention the problem of appropriately defining the CS, but their annotation is restricted to the sentence-level only.

There are many different types of superlatives, either through the way they are expressed in syntax (e.g. adverbial vs. adjectival) or through the way they express a semantic comparison. But most works only focus on single subtypes and do not cover \textit{all} forms of superlatives.

In this paper we propose a unified annotation schema for providing a complete semantic reading of superlatives. The schema defines the {\em superlative frames}, encapsulating all the elements needed for  an interpretation. The frames remain identical whether or not the semantic elements are explicit or implicit, and allow us to specify restrictions from context, which are made explicit in the form of Neo-Davidsonian Semantics, allowing one to show when interpretations are restricted by events or arguments. 

Based on the proposed schema we annotate a dataset of superlatives, called \dataset{}\footnote{https://github.com/ValentinaPy/SuperSem},
over different domains, ranging from encyclopedic text to dialogue. We show that \dataset{} contains a large variety of superlative types and interesting instances of implicit domain restrictions.

Given \dataset{}, we train models based on T5 \cite{raffel2020exploring} to predict superlative interpretations. This allows us to analyze the effect the discourse context has on restricting possible comparison interpretations and to assess the efficacy of filling in implicit elements from larger contexts. We also show that it is challenging for GPT-4 to appropriately incorporate discourse restrictions for the interpretation of superlatives.

Superlatives are most interesting because they, on the one hand, involve deep semantic understanding and require inference of implicit elements in text, while on the other hand are highly prevalent in many downstream applications.

\section{The Challenge: Syntax, Semantics and Pragmatics of Superlatives}
The particular challenge of superlatives interpretation is somewhat ignored in the study of natural language understanding,  despite demonstrating many interesting syntactic and semantic phenomena. While all superlative expressions seem to do the same thing, i.e. pick a maximal entity/event, they appear in different forms, which, for NLP, hinders their uniform interpretation. Furthermore, superlatives may appear in various syntactic realizations which do not explicitly express some of the semantic aspects of the comparison, in particular how the comparison set is being restricted by context. 
In what follows we describe the possible syntactic realizations and semantics of superlatives, and how some of the frames can be implicit or ambiguous.

\subsection{The Syntax of Superlatives}
Humans use superlatives in order to reason about quantities and degrees in a comparative manner. Analytically, in English, they are formed using the adverbs \textit{most} and \textit{least} and inflectionally they are formed with the addition of the suffix \textit{-est}. The comparison can either be performed in a positive (\textit{most}), or negative orientation (\textit{least}, \textit{few}) \cite{english.grammar.2002}.

Superlatives can be broadly categorized into the following superficial forms: adjectival superlatives, such as ``Mia is the tallest girl''; adverbial superlatives, such as ``Most commonly, psychologists use surveys''; and other forms which are not superlatives morphologically, but still lexicalize a superlative meaning, i.e. ``The main reason'' \cite{scheible2009computational}.

\subsection{The Semantics of Superlatives}
\label{sec:semantic_description}
Semantically, superlatives perform a domain based comparison \cite{szabolcsi1986comparative, alshawi1992core, gawron1995comparatives, heim1999notes, farkas2000comparative}. 

\ex. \label{secondex}\small{ Nemo is the smallest fish out of all the fish in the aquarium.}

In Example~\ref{secondex} the \textbf{target} of the comparison  (i.e., the element that has the max/min of some set) is \textit{Nemo}. The \textbf{target} is being compared to a set of other entities, the \textbf{comparison set}, which in this example is defined by \textit{all the fish in the aquarium}. Each item in the \textbf{comparison set} has a \textbf{property} along which it is being compared, in this example the \textbf{property} is \textit{size}. As we seek the {\em smallest} fish, the \textbf{orientation} of the size comparison is negative. By inference, comparatives can convey the same sense as superlatives, e.g. ``Nemo is smaller than any of the other fish.''

Towards a semantic account of superlatives, \citet{scheible2012textwiki} defined 3 types of superlative comparisons. The \textbf{Property Set Comparison} is the most known superlative type, where members of the \cs{} are being compared with respect to the \textit{property}. Ex.~(2) illustrates a property set comparison, where all fish in the aquarium (CS) are being compared by the \textit{size} property.
The \textbf{Relative Set Comparison} involves two interdependent set comparisons. For example, in ``Of all the band members, Bob played the longest solo.'', the set of `solos' is being compared in terms of \textit{length}. But this set is further restricted by a second set, the `band members playing (solos)'.
The \textbf{Subject-based Set Comparison} is peculiar in that the \cs{} does not consist of different entities, but instead compares the \target{} at different \textit{states}: ``Bob is \textbf{hungriest} at noon.''
Here the comparison set involves Bob's level of hungriness at different times of the day.

\subsection{Implicit Elements of Superlatives}
\label{sec:implicit}
Superlatives can appear in various syntactic realizations which do not explicitly express some of the aspects of the comparison.
Often an explicit \target{} is missing:

\ex. \label{extarget} \small{ a. \underline{Nemo} is the \textbf{smallest} \textit{shark}.
\\b. The \textbf{smallest} \textit{shark} hides under a rock.}

In  Example \ref{extarget}a. the \target{} 'Nemo' is explicit, while in \ref{extarget}b. the superlative NP  stands for the  implicit  \target{}. 
The \cs{} (and its domain restrictions) can also be (entirely) missing:
 \ex. \label{excs} \small{ a. \textit{In Europe}, \underline{he} is the \textbf{tallest} \textit{man}.\\ b. \underline{He} is the \textbf{tallest}.}
 
In \ref{excs}b. the head of the superlative NP (''man'') is empty \cite{elazar2019s} and the domain ''In Europe'' is implicit, while in \ref{excs}a. constructing the CS consists of two, syntactically non-adjacent spans, i.e., 
\textit{men}, and \textit{in Europe}.

Even if the CS's head is not missing, there might still be restrictions from the broader discourse to be found:
 \ex. \label{exdiscourse} \small{For years, many Haitians and their descendants \textit{in Cuba} did not identify themselves as such [...].}
After Spanish, \underline{Creole} is the second \textbf{most-spoken} \textit{language}.

The complete \cs{} in \ref{exdiscourse} is `languages spoken in Cuba', which could only be identified by also including the previous context. 

The importance of the domain of interpretation for disambiguating various superlative readings has also been pointed out in formal semantic literature \cite{gutierrez2006superlative}.
 \ex. \label{expresent} \small{She gave me the \textbf{most expensive} \textit{present}.}
 
Without context this example has multiple readings (absolute vs. relative \cite{szabolcsi1986comparative, heim1999notes, farkas2000comparative, huddleston2005cambridge}): The CS could be `presents in the world' (absolute) or `presents I have \textbf{received} from my friends on my birthday' (relative) or `presents she \textbf{gave} me on that day' (relative), etc. Note the restricting events in the last two interpretations, making them relative set comparisons. 

Lastly, the subject-based set comparison (Sec.~\ref{sec:semantic_description}) is very implicit as neither the \target{} nor the \cs{} are explicitly expressed in syntax:
 \ex. \label{exdsubject} \small{The human is \textbf{broadest} at the shoulders.}
 
 Here the implicit target would be `the width of a human at the shoulders' and the implicit comparison set would be `the width of different parts of the human body'. This illustrates how critical it is for machine comprehension to infer implicit elements in order to retrieve the correct entities.


\subsection{Context Restrictions}
The discourse context which propositions appear in has an influence on the interpretation of such propositions, which can broadly be described as the problem of context dependence. A specific case of context dependence is quantifier domain restriction \cite{geurts1999domain, stanley2000quantifier}, which includes the case of superlatives \cite{gutierrez2006superlative}.

\ex. \label{exrest} \textbf{Every} cupcake is gluten free.

Assuming that Bob brought cupcakes to a birthday party and utters the sentence in \ref{exrest}, it is clear that not all cupcakes in the whole universe are being referred to and quite probably also not all cupcakes at the party. Rather, the expression is equivalent to ``Every cupcake Bob brought to the party is gluten free.'', which can be inferred from context. 

Besides the case of quantifiers, context dependence also occurs with pronouns or demonstratives, such as ``\textbf{This} book is really good.'', 
where the referent can only be understood through additional context (e.g. by seeing what someone is pointing to).
Ambiguous propositions (either lexically or structurally) can also be disambiguated with context \cite{stanley2000quantifier}. But also non-ambiguous sentences could have a positive or negative truth value depending on context. And lastly, contexts also have pragmatic functions, for example by marking irony.

In this paper we study how context affects and restricts superlative readings, with a focus on superlatives where parts of the comparison are left un- or under-specified.

\section{Superlative Frames}
\label{sec:frames}
\begin{figure*}[ht]
    \centering
    \includegraphics[width=\textwidth]{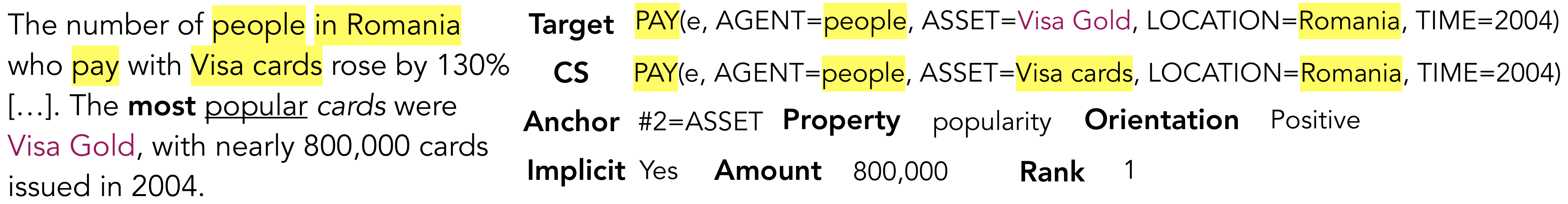}
    \caption{An annotation example showing on the left the superlative (most), the sentence it appears in, and its previous context (shortened). On the right it shows the annotation slots (Target, DOI, CS etc.) and how they are filled given the text. Highlighted in yellow are the implicit discourse restrictions.
    }
    \label{fig:anno_ex}
\end{figure*}
In what follows we define the set of superlative frames. We propose a formal, event-based, account of superlatives.
We first define all the frames and then show how they are able to cover all superlative types from Sec.~\ref{sec:semantic_description}. These superlative frames provide an intuitive way to achieve (i) high quality annotations (see Sec.~\ref{sec:annotation}) and (ii) a straightforward use for computational modeling.

The frames are built with a focus on annotating the semantics of the comparison and making discourse restrictions explicit. Additionally, we center the frames around events/predicates, when available, using Neo-Davidsonian semantics. This is motivated by the fact that events can also function as set restrictions for superlatives, such as for \textit{relative set comparisons}. When no event is restricting the superlative, we annotate the restricting noun phrases.
An example annotation using our scheme can be seen in Fig.~\ref{fig:anno_ex}.

\paragraph{Comparison Set}
In the \textit{comparison set} we define the set of entities or events that take part in the comparison: $CS=\{e_1, ..., e_n\}$.
Comparisons involving an event are formulated using a neo-davidsonian expression. The argument slots are labeled using VerbNet roles \cite{schuler2005verbnet} and filled with tokens from context. The \cs{} in Fig.~\ref{fig:anno_ex} consists of a \textit{pay} event, with four semantic arguments: \textsc{agent}, \textsc{asset}, \textsc{location} and \textsc{time}.

\paragraph{Property}
Each entity or event in the \textit{comparison set} has a property along which it is being compared. We use nouns to define these properties. The property in the example is \textit{popularity}.

\paragraph{Target}
The \textit{target} stands in an IS-A relation with the \textit{comparison set}, i.e. the target is one of the entities or events in the \textit{comparison set}: $t \in CS$. Specifically, it is the entity or event whose property has the max/min value: $max/min(p)$. 

\paragraph{Anchor}
The \textit{anchor} of the \cs{} designates the focus of the comparison. We index its position in the CS, e.g. \#2=ASSET. The \cs{}, expressed in words, would be something like `Visa cards people pay with in Romania'. The \textit{anchor} signals that we are comparing `Visa cards' and not another entity. 

\paragraph{Orientation +/--}
This field designates if the min or max operation was applied on the property.

\paragraph{Rank}
Sometimes superlative targets do not denote the entity at the min/max position, but instead they denote an entity at the n-th position. For example: ``the \textit{second} biggest Bulgarian port''. In these cases we note the given rank (default is 1).

\paragraph{Implicit +/--}
This field specifies whether the superlative is restricted by content outside the sentence boundary or alternatively by content that is not mentioned but implied. 

\paragraph{Amount}
The \textit{amount} specifies the realization of the \textit{property}.
In Fig.~\ref{fig:anno_ex} it is explicitly mentioned that the amount of `800,000 cards issued' makes the `Visa Gold' card the most popular one.

\section{Annotating Superlatives}
\label{sec:annotation}
One of our contributions is the \dataset{} dataset, consisting of more than 4000 annotations of superlatives and their semantic interpretation in terms of the set of frames described in Sec.~\ref{sec:frames}. In what follows, we describe the annotation process and provide an analysis of the final dataset itself.

\subsection{Data}
In order to cover a variety of domains we annotate the following datasets: We have re-annotated the Superlatives Wikipedia corpus \cite{scheible2008annotating}, two dialogue datasets: Dailydialog \cite{li2017dailydialog}, MultiWOZ 2.2 \cite{zang2020multiwoz}, a subset of superlatives in Amazon Product Reviews \cite{ni2019justifying}, superlatives found in the Wikinews documents used by TNE \cite{elazar2022text} and  superlatives in passages from the following narrative texts: \textit{Animal Farm} by George Orwell, \textit{Harry Potter and the Philosopher's Stone} by J. K. Rowling, \textit{The Hitchhikers Guide to the Galaxy} by Douglas Adam, \textit{The Great Gatsby} by F.~Scott Fitzgerald and \textit{The Hobbit} by J. R. R. Tolkien.

In order to extract sentences containing superlatives from our chosen corpora, we POS-tag them using Stanza \cite{qi2020stanza} and extract all sentences containing at least one word tagged with either JJS (adjectival) or RBS (adverbial). We run the preselection by POS-tag approach on a test set from \citet{scheible2008annotating}, receiving a recall of 98.8\%. We can therefore be sure that we are capturing most, if not all, superlatives present in a text. In terms of precision, we note that not all words tagged with JJS signal a superlative, such as, `at least' or `at most', which are proportional quantifiers. We increase precision to 99\% through manual post-processing, by having annotators mark such instances as non-superlative readings.

\subsection{Annotators}
We hired two annotators for the task, who were provided with guidelines and training sessions. For quality assurance the authors met with the annotators in weekly meetings and discussed a subset of the annotations. Additionally, we periodically calculated IAA between the annotators and an expert (i.e. author of the paper, from here on called annotator C). The annotators were paid above minimum wage for the region. One of the annotators was a Master's student in linguistics, hereafter called annotator A, and the other, annotator B, a Bachelor's student in Computer Science.

\subsection{Coverage}
\label{sec:coverage}
The superlative frames, described in Sec.~\ref{sec:frames}, can be used to annotate all three semantic types defined by \citet{scheible2012textwiki}. The example in Fig.~\ref{fig:anno_ex} annotates a \textit{relative set comparison}. A \textit{property set comparison} distinguishes itself from the \textit{relative} one by not having restrictions, which can either be events or noun phrases, in the \target{} and \cs{}. The \textit{subject-based set comparison} can be identified through our annotations by the use of light verbs as the event predicate. For ``Bob is hungriest at noon'', the \target{}, for example, would look as follows: BE\_HUNGRY(e, THEME=Bob, TIME=at noon).
\begin{figure}[ht]
    \centering
    \includegraphics[width=0.45\textwidth]{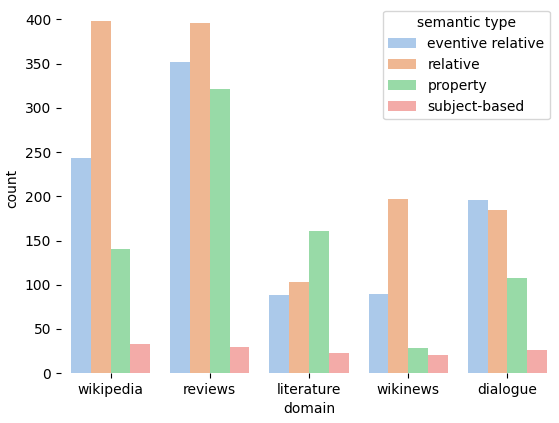}
    \caption{Counts of the semantic types annotated over the different domains.}
    \label{fig:semantic_coverage}
\end{figure}

Figure~\ref{fig:semantic_coverage} shows how the types are distributed across the five domains. Except for the literature domain, relative superlative comparisons are the most frequent. In the literature domain on the other hand, property set comparisons occur most often. The least frequent type is the subject-based comparison, for all domains.

\subsection{Dataset Analysis}
Here we present an analysis of the \dataset{} dataset and report statistics.
Table \ref{tab:data_stats} shows the general dataset counts, split by domain. The final dataset consists of more than 3k annotated superlatives and more than 1k non-superlatives, which are pos-tagged as JJS, but do not express a superlative reading (such as `most' being used as a quantifier). About 42\% of annotated superlatives contain implicit elements and about 35\% contain an event.
\begin{table}[ht]
\small
\begin{tabular}{l|llll}
Domain         & Sup. & ¬Sup. & Events & Implicit \\ \hline
Wikipedia      &  814    &   476       &    274       &  242      \\
Reviews        &  1098   &   286       &    363       &  555      \\
Dialogue       &  522    &   219       &    222       &  293      \\
Literature     &  376    &   186       &    111       &  92      \\
Wikinews       &  336    &   152       &    109       &  146      \\ \hline
total          &  3146   &   1319      &    1079      &  1328      \\
\end{tabular}
\caption{Dataset counts split by domain, showing how many superlatives (Sup.) or non-superlative (¬Sup.) there are in the dataset. We further show numbers on how many superlatives were marked as being implicit and how many superlatives were restricted by events.}
\label{tab:data_stats}
\end{table}

\paragraph{Events}
Overall, the events restricting the CS are diverse, with 353 distinct predicate lemmas. The most common predicates include \textit{have}, \textit{do}, \textit{use}, \textit{find} and \textit{make}. Light verbs are frequent because they are used to express \textit{subject-based set comparisons} (Sec.~\ref{sec:coverage}). Other common verbs, which are not light verbs, are \textit{create}, \textit{play}, \textit{own} and \textit{buy}.

\begin{figure}[ht]
    \centering
    \includegraphics[width=0.45\textwidth]{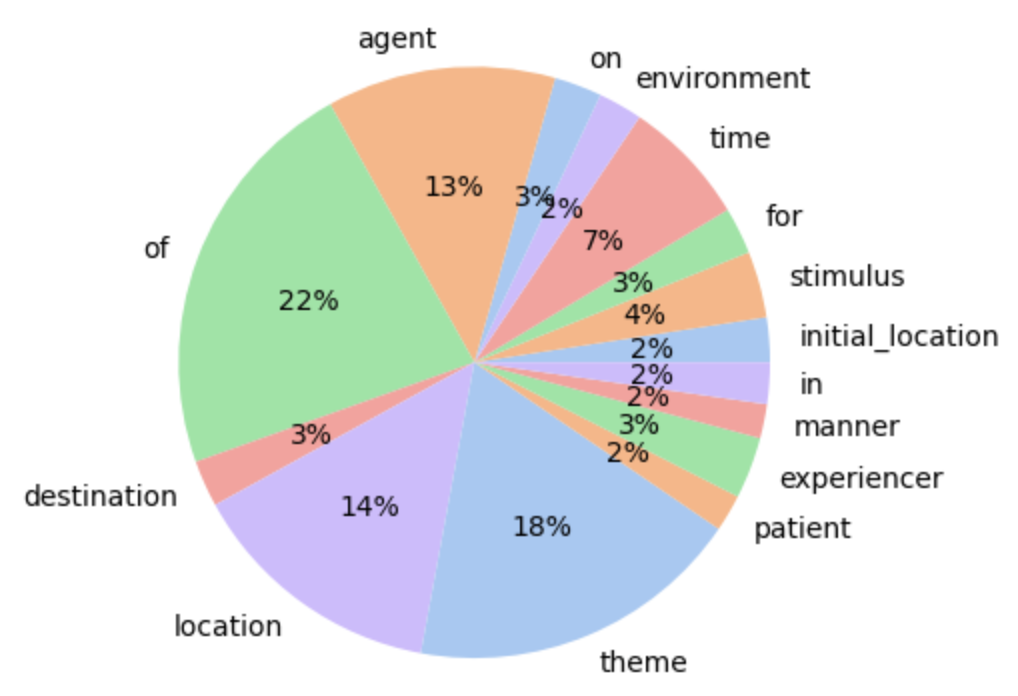}
    \caption{Most frequent roles found in \dataset{}.}
    \label{fig:arg_analysis}
\end{figure}

\paragraph{Arguments}
In Figure \ref{fig:arg_analysis} we visualize the distribution of the most frequent roles in our \target{} and \cs{} annotations. \textsc{Agent}, \textsc{location} and \textsc{theme} are the most frequently annotated VerbNet roles. We additionally allowed the use of `of' as a slot designating restricting bridging relations (such as ``writers OF=the ancient world'').

\begin{figure}[ht]
    \centering
    \includegraphics[width=0.45\textwidth]{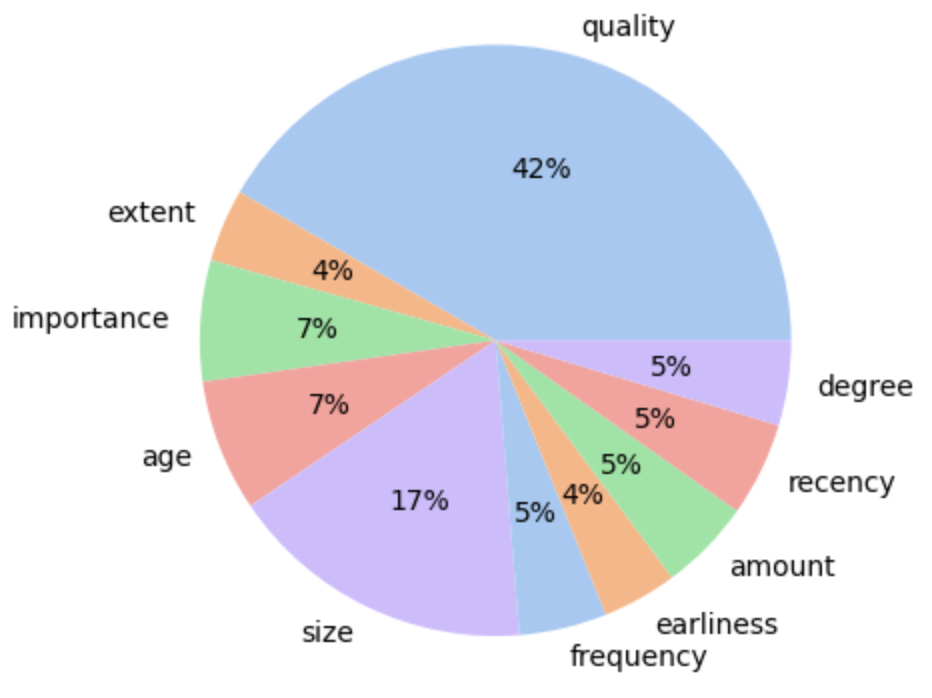}
    \caption{Most frequent properties in \dataset{}.}
    \label{fig:property_analysis}
\end{figure}

\paragraph{Property Types}
In Figure \ref{fig:property_analysis} we visualize the distribution of the most frequent \textit{properties}. \textit{Quality} (`best'/`worst') and \textit{size} (`biggest'/`smallest') are most prevalent in the dataset.

\subsubsection{IAA}
\label{sec:IAA}
To ensure annotation consistency and quality, we performed three rounds of IAA checks. For the first two checks, a set of instances was given to two annotators: 50 Amazon review instances were shown to annotator A and C, and 81 Wikipedia instances to annotators B and C.  We compare the agreement on different aspects of the annotation, as seen in Tab.~\ref{tab:IAA}. For all categorical values we also report Cohen's Kappa scores, which are moderate to high for the \textit{event vs. no-event} and the \textit{orientation} frames and fair for the \textit{implicit} frame. It is worth noting that these first IAA checks were performed while annotations were on-going and that there was an annotation consolidation after the checks. This means that the scores could be seen as a lower bound for annotation agreement, which then improved after the consolidations.
Higher exact match agreement was harder to achieve for some non-categorical categories, like the \cs{}. This is mainly due to the order in which arguments are listed and the differences in argument spans (i.e. determiners being included or excluded) and lower agreement for these categories does not necessarily indicate wrong annotation.

The third IAA check was performed after the annotation effort was completed. We randomly sampled 30 instances from the whole dataset, which were then re-annotated by annotator C. The results show that agreement has improved for nearly all categories. This is a promising sign for our annotation protocol, since it indicates that our annotator training and consolidation process resulted in higher agreement.

\begin{table}[ht]
\small
\begin{tabular}{l|lll}
                                            & \multicolumn{1}{l}{Wiki.} & \multicolumn{1}{l}{Reviews} & Final \\ \hline
event vs. none                          & .78 (.55)                         & .76        (.47)  & 0.83 (.63)              \\
exact target                                & .23                          & .29  & .4                      \\
exact CS                                   & .1                           & .29   & .33                     \\
exact anchor                                & .58                          & .45  & .5                      \\
exact property                              & .58                          & .61  &  .77                     \\
exact orientation                           & .99    (.88)                        & .92        (.86)       & 1 (1)           \\
exact implicit                              & .67        (.34)                    & .43       (.16)        & .73 (.48)           \\
event predicate                  & .78                          & .73                       & .75 \\
CS (no event)                      & .22                          & .54     &                 .47  \\
role arg. iou\textgreater{}=0.5 & .36                          & .42   &  .45                  
\end{tabular}
    \caption{Inter-Annotator Agreement on Wikipedia, Reviews and a final set randomly sampled from \dataset{}. \textit{event vs. none}: Accuracy of choosing an event. \textit{exact ...}: exact match accuracy for each of these slots. \textit{event predicate}: acc. of choosing the same predicate. \textit{CS text}: exact match of CS text if no event was chosen. \textit{role argument}: Accuracy of having an intersection over union (IOU) \textgreater{}= 0.5 of the role argument text. Cohen's Kappa scores are added in brackets for all categorical values.}
    \label{tab:IAA}
    \end{table}

\subsubsection{Bridging and Discourse Restrictions}
Discourse restrictions and bridging relations are closely related. Bridging relations are anaphoric relations, such as part-of, between two entities in a discourse \cite{gundel1993cognitive, hou2013global}. They form a subset of the different types of discourse restrictions affecting the \cs{} of superlatives. To analyze the influence of noun phrase (NP) relations as discourse restrictors for superlatives, we annotated a subset of TNE \cite{elazar2022text}. TNE annotations follow a broader definition of bridging, by connecting NPs of any relation type. 

\ex. \label{tneex} \small{The \textbf{largest} single language is English, which has 2.3 million articles.}

The sentence in Ex.~\ref{tneex} has the \textit{target} `English' and the \cs{} `single languages'. The \cs{} is further restricted by context: `single languages \textbf{OF=Wikipedia}'.
This type of NP part-of relation is also annotated in TNE, where `largest single language' is connected with the preposition `of' to `Wikipedia editions'.

Using string matching we found that \textbf{about 26\% of the implicit superlatives in our TNE subset are restricted by noun phrase relations} also present in TNE. Due to the automatic way of extracting these statistics we assume that this is a lower bound and conclude that CS are often also restricted by NP relations from discourse.

\subsubsection{Implicit Arguments as Discourse Restriction}
Since events can also take part in superlative comparisons, implicit arguments also form a subgroup of discourse restrictions. Implicit arguments \cite{ruppenhofer2010semeval, gerber2010beyond, roth2015inducing} fill semantic roles of predicates, where the argument is not syntactically connected to the predicate and might even be found outside of the predicate's sentence. 

\ex. \label{implicitsrlex} \small{\textbf{Most commonly}, psychologists use paper-and-pencil surveys.}

Ex.~\ref{implicitsrlex} contains the verbal predicate `use'. In VerbNet `use' has 3 possible roles, of which 2 are explicitly filled in ex.~\ref{implicitsrlex}: the \textsc{agent}, with `psychologists' and the \textsc{theme}, with `paper-and-pencil surveys'. The third role, \textsc{eventuality}, is implicit and can be filled from the previous context, with `observational studies', restricting the CS as follows: USE(e, AGENT=psychologists,
THEME=surveys, EVENTUALITY=observational studies)

This could be paraphrased as: \textit{out of all types of surveys psychologists use for observational studies}. We find, with automatic string matching of argument text and context, that \textbf{about 67\% of event-restricted superlative instances have one or more implicit argument from context}.

\section{Computational Modeling of Superlative Semantics}
In what follows we want to examine the computational modeling of superlative semantics. We are interested in multiple aspects. First, we want to establish sequence-to-sequence baselines for predicting our superlative frames, when trained on \dataset{}. Additionally, we want to better understand the role of context when predicting the superlative frames, and when superlatives are ambiguous. 
To answer those questions, we carry out 5 types of experiments: In the first experiment we learn to predict the semantic interpretation of superlatives using seq2seq models. In the second experiment we predict discourse context given the formal semantic interpretation, in order to see whether the model is able to appropriately identify context restrictions. In the third experiment we test GPT-4's capabilities of identifying the comparison set. As a fourth experiment, we test our model on a hand-crafted challenge set of ambiguous superlative expressions. And lastly, we quantify the ambiguity of superlative expressions and how context can restrict the set of possible interpretations.

\paragraph{Experimental Setup}
We use T5-3B \cite{raffel2020exploring} for all experiments, if not noted otherwise. We use a batch size of 2, a maximum output length of 300 and we train for 3 epochs on 2 A100 GPUs. For training, development and testing we create 80-10-10 splits of \dataset{}, by randomly sampling from each domain equally.

\subsection{Predicting the Interpretation}
\label{sec:prediction}
We first want to establish a seq2seq baseline for predicting superlatives frames, trained on \dataset{}. Given as input a context with a superlative expression we want the model to predict the appropriate superlative interpretation, by filling the frames, such as \target, \cs{} and \textit{property}. We experiment with different input/output settings. 
\begin{enumerate}[noitemsep]
    \item \textsc{full}: predict all the frames at once.
    \item \textsc{single}: fine-tune a model for each slot individually. 
\end{enumerate}

To see the effect of context on predicting frames, we either only use the single sentence the superlative occurs in, or use the full context.

\paragraph{Results}
Table \ref{tab:bigtab} displays the results. For most of the frames, the setting which includes both the superlative sentence and the additional discourse context works best. This indicates that the \textbf{context contains further information needed to make the appropriate inferences for predicting superlative interpretations}. We also see that, except for the \textit{property} slot, training a \textbf{specialized model for each slot works better} than training a general model that predicts the full annotation at once. The results on \textbf{eventive superlatives only (in grey), show that they form a specially challenging subset for models}. 
The best models still do not achieve the same EM scores as the human IAA scores (Sec.~\ref{sec:IAA}). While these IAA scores come from a different, and smaller, set than the test set, they can still provide a reference point.

\begin{table}[]
\small
\begin{tabular}{l|llllll|l}
                                                                        & \multicolumn{3}{l}{Sentence}            & \multicolumn{3}{l}{Sent. + Context}    & \multicolumn{1}{l}{Hu.}         \\
                                                                        & EM          & IOU         & R           & EM          & IOU         & R           & EM \\ \cline{1-8}
\textbf{target}                                                                  & 24          & 46          & 67          & 29          & \textbf{53} & \textbf{73} &  40     \\
\textsc{full}                                                             & 26          & 46          & 65          & \textbf{30} & 51          & 69          &     \\
\rowcolor{Gray}  \textsc{event} & 8           & 34          & 66          & 9           & 39          & 70          &       \\
\textbf{CS}                                                                      & 25          & 43          & 68          & \textbf{31} & \textbf{48} & \textbf{72} &  33     \\
\textsc{full}                                       & 21          & 32          & 51          & 22          & 33          & 55          &     \\
\rowcolor{Gray} \textsc{event}     & 13          & 31          & 65          & 18          & 40          & 71          &       \\
\textbf{anchor}                                                                 & 50          & 53          & 61          & \textbf{58} & \textbf{60} & \textbf{67} &   50    \\
 \textsc{full}                                   & 36          & 40          & 51          & 42          & 47          & 56          &     \\ \cline{1-8}
\textbf{prop.}                                                                & \textbf{72} & \textbf{73} & \textbf{75} & 70          & 71          & 73          &  77     \\
\textsc{full}                                  & 71          & 71          & 73          & 71          & 71          & 73          &     \\
 \cline{1-8}
\textbf{orient.}                                                                 & \textbf{92} & 92          & 92          & 87          & 87          & 88          &   100    \\
 \textsc{full}                                  & \textbf{92} & 92          & 92          & 91          & \textbf{93} & \textbf{93} &    \\
 \cline{1-8}
\textbf{impl.}                                                                & 69          & 69          & 69          & \textbf{73} & \textbf{73} & \textbf{73} & 73    \\ \cline{1-8}
\end{tabular}
    \caption{Results showing exact match accuracy (EM), intersection over union (IOU) and Rouge-n (R, n=1). For each semantic slot we show the performance when only trained on that specific slot and the \textsc{full} performance. For \textit{target} and \cs{}, highlighted in grey, we also show performance of the \textsc{single} model (\textsc{event}) tested only on the events subset. The last column shows human EM IAA scores from the last IAA round.}
    \label{tab:bigtab}
\end{table}
\subsection{Frame-specified Context Generation}
We also model the problem in the inverse direction: given the semantic interpretation, predict further restricting context outside of the sentence boundaries.
Specifically, we trained a model on the task of predicting the whole paragraph given the superlative interpretation and the sentence the superlative appears in.

This type of context generation is challenging because it requires the model to perform semantic consolidations \cite{hirsch2023revisiting}: it needs to identify propositions which are expressed in the superlative frames annotations, but not in the given sentence, and then generate coherent and appropriately restricting context.

\paragraph{Results}
The context generation model achieved a decent ROUGE-1 score of 0.41, which indicates that the model learns to generate appropriate context restrictions.
We further performed a manual evaluation of the test set results, where we analyzed how well the model is able to predict context that appropriately restricts implicit superlative readings. For example:

\ex. \label{context} \small{But the ancient race of the northern mountains were the \textbf{greatest} of all birds [...]. \textbf{TARGET:} Eagles LOCATION=northern mountains \textbf{CS:} birds \textbf{ANCHOR:} birds \textbf{PROPERTY:} greatness \textbf{ORIENTATION:} positive}

Here, the \target{} is implicit. The model is nonetheless able to identify which parts of the superlative interpretation are not mentioned in the given sentence and then predicts appropriate context containing this implicit information. In this case, it predicted context mentioning \textit{eagles}\footnote{`Then the eagles swooped down and snatched him up, and he flew away [...]''}. \textbf{In 83\% of the implicit test cases, the model correctly included the missing restrictions when predicting context beyond the sentence boundaries}.

\subsection{Superlatives and GPT-4}
We test GPT-4\footnote{Accessed on: 10.01.2023 and 05.16.2024.}'s ability to zero-shot interpret superlative comparisons, in natural language, and to few-shot interpret superlatives' \cs{}.

\paragraph{Experimental Setup}
We perform two different experiments on the test split of \dataset. First, a zero-shot experiment where we input a single sentence containing a superlative expression and ask GPT-4 to answer the question ``What is being compared to what here with the superlative?''. The model's answer is expected to be a natural language explanation of the comparison. We also experiment with adding the full context and with explicitly marking the superlative in the prompt.
Second, we evaluate GPT4 in a few-shot manner, where the task is to predict the superlative frame of the \cs{}. Specifically, we add three demonstrations to the prompt, with each demonstration capturing a different type of \cs{} interpretation.

For the first setting, the NL explanation is evaluated with human evaluation, as it differs from the logical forms contained in \dataset{}. 
And for the few-shot setting, we additionally evaluate using the same metrics we also used to evaluate the fine-tuned T5 models (Sec.~\ref{sec:prediction}).

\paragraph{GPT-4 does not always recognize that there is a superlative comparison.}
In the first setup, our aim was to see whether the LLM is able to recognize the comparison relation triggered by the superlative. For 16\% of the test set sentences GPT-4 either outputs that ``There is no direct comparison being made in this sentence.'', or, in rarer cases, mentions other types of comparisons present in the sentence, such as discourse relations. Generally, though, it is able to correctly recognize that the presence of a superlative expression indicates a comparison.

\paragraph{Discourse restrictions make things harder.}
\begin{table}[ht]
\small
\begin{tabular}{l|ll}
                                & target & CS \\ \hline
single sentence  (implicit+explicit)               & 89.0   & 77.9           \\
paragraph (implicit)       & 84.0   & 62.9           \\
single sentence (implicit) & 87.1   & 69.7    \\ \hline
paragraph (implicit) few-shot &  -   &  32.6     \\

\end{tabular}
    \caption{GPT-4's performance (accuracy) on identifying the target and CS, evaluated on the single sentence and the paragraph level.}
    \label{tab:gpt}
\end{table}

In Tab.~\ref{tab:gpt} we show the target and \cs{} accuracies for the single sentence context, the paragraph-level context for the implicit subset and the single sentence setting for the implicit subset.
The main conclusion to be drawn from the results is that the paragraph-level interpretation of superlatives is harder than the single sentence setting, for GPT-4. The following is an example of a failure case:

\ex. \label{drfailure} \small{The Four Horsemen: Book 2 in the Light Trilogy was intense. [...] I think out of all of the characters, excluding the main ones, I would have to say that I love Mona the \textbf{most}. [...]}

In this excerpt (shortened for space considerations), the \target{} is `Mona' and the \cs{} is `all of the characters, excluding the main ones', which is further restricted by a \textit{love} event and by the fact that these characters are from the book `The Four Horsemen'. GPT-4 writes: ``All other characters are being compared to Mona with the superlative.'' While this output correctly identifies that there is a comparison between Mona and other characters, it incorrectly writes `all other characters'. The correct response would have excluded the main characters from the comparison. It further misses to specify discourse restrictions, such as the book title these characters appear in.

\paragraph{Few-shot is not enough for learning about superlative semantics.}
The few-shot experiments show that structured semantic prediction is hard to do using prompting. The few-shot scores in Tab~\ref{tab:gpt-few-shot} fall behind the fine-tuned model at predicting the \cs{}. These results are in line with recent works examining prompting for structured prediction: \citet{ettinger2023you} found that LLMs are limited in their capability to predict correct AMR structures, also when using few-shot demonstrations, and \citet{mehta2024promptly} showed that prompting for semantic structures leads to inconsistencies.

In addition, the manual evaluation reveals lower accuracy for the few-shot prompting setup (Tab.~\ref{tab:gpt}). Looking at the outputs, the model sometimes seems to be able to capture the format of the frames, also of eventive CSs. Interestingly, most of the errors involve the model not being able to incorporate the relevant elements from context, such as missing a LOCATION or TIME restriction.

\begin{table}[ht]
\small
\begin{tabular}{l|lll}
     & EM & IOU & R \\ \hline
T5 fine-tuned   & 31 & 48  & 72 \\
GPT4-few shot & 4  & 17  & 43  \\
\end{tabular}
    \caption{GPT-4's few-shot performance (EM - exact match, IOU - intersection over union, R - Rouge-1) on full test set, for identifying the CS, given full context.}
    \label{tab:gpt-few-shot}
\end{table}

\subsection{Ambiguous Superlatives}
\label{sec:ambig}

\begin{figure}[ht]
    \centering
    \includegraphics[width=0.45\textwidth]{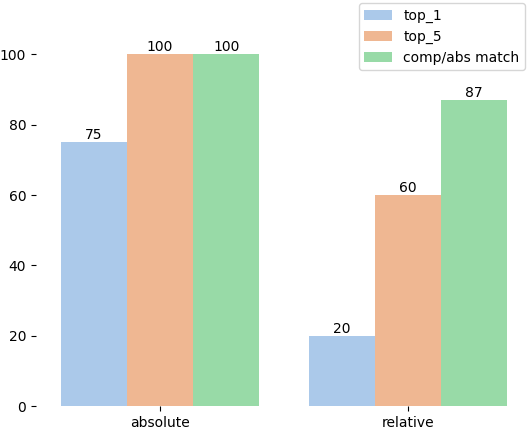}
    \caption{Accuracy for predicting the CS: given \textit{absolute} vs. \textit{relative} contexts. \textit{top\_1}: the first prediction in a beam is correct. \textit{top\_5}: at least one prediction in a beam of 5 is correct. \textit{comp/abs match}: Does the type (absolute/relative) of the predicted CS fit the gold type?}
    \label{fig:ambig_fig}
\end{figure}

The main ambiguity present for superlatives is the distinction between absolute versus relative interpretations (see Sec.~\ref{sec:implicit}). In an effort to analyze how sensitive our model is to discourse cues which could help to disambiguate between different readings, we perform the following experiment. 
We manually curated 20 sentences in which the superlative comparison is ambiguous. Many of these sentences are (synthetic) example sentences found in formal semantic literature. Additionally, for each of the 20 instances, we added context that strengthens a certain reading, such as absolute vs. relative readings. In what follows we show an example of our synthetic test set.
\ex. \label{plant}
\small{John put the \textbf{tallest} plant on the table.\\
Context 1: \textit{Tom, John and Mary all brought plants which they put on the table.}}

The first sentence is ambiguous in that it could be read as either absolute or relative, i.e. restricted by the \textit{putting} event or not. Given the additional context 1, the relative reading is strengthened: PUT(e, AGENT=Tom \& John \& Mary, PATIENT=plants, DESTINATION=table). 

We run the T5-3b model trained to predict the \cs{} slot on the synthetic test data. 

\paragraph{Relative is harder than absolute} As shown in Fig.~\ref{fig:ambig_fig}, absolute superlative comparisons are easier to identify: in 100\% of absolute cases the predicted \cs{} represents an absolute reading. Additionally, in 100\% of the absolute test cases, the model predicts the correct \cs{} among a beam of 5. Relative readings, on the other hand, are harder to get right and the model also only correctly identifies a relative instance as such in 87\% of the cases.

\subsection{Ambiguity and Context}

To analyze and quantify ambiguity and the effect of context in superlatives, we look at two measures, \textit{conditional log-probabilities} and \textit{entropy}.

Formally, for the \textit{conditional log-probabilities}, we define the \textit{prefix} to be the previous context and the \textit{stimulus} to be the superlative-sentence, consisting of tokens $W_n=(w_1, ..., w_n)$. And we then calculate the conditional per-token log-probabilities, using \textsc{minicons} \cite{misra2022minicons}: 
\[\frac{\sum_{n=1}^{|W|}p(w_n|\mbox{prefix})}{|W|}\]

With \textit{entropy} we aim to measure how models deal with ambiguous superlatives and whether they are able to express all possible interpretations of such instances. For this purpose we quantify the entropy of interpretation types present in a top-n beam search output:
\[H(X) = -\sum_{x \in \mathcal{X}}p(x)\log p(x)\]
Where $X$ takes values from the following interpretation types: eventive relative set comparison (SC), (non-eventive) relative SC, property SC and subject-based SC. 

We evaluate the output of our T5-3B model, trained to predict the \cs{}, on the test split of \dataset{} and the synthetic challenge set.

\subsubsection{Probability Given Context}
With the \textit{conditional log-probabilities} we want to measure the likelihood our model assigns to different given interpretations of the synthetic inputs with varying contexts. Concretely, we take a superlative sentence and see how the likelihood of an interpretation changes given different tailored contexts, or no context. 

Overall, our model prefers the correct over the incorrect interpretation in 87\% of the cases. Interestingly, the absolute difference between the log probabilites of two completions given only a sentence, is on average smaller than the absolute difference between the log probabilities of two completions given the full context. \textbf{This indicates that a model fine-tuned on \dataset{} is appropriately sensitive to ambiguous instances without context}, i.e. assigns all possible completions similar likelihoods, while given the full context, the likelihood gap increases and a certain interpretation becomes considerably more likely.

\ex. \label{party}
\small{John is angriest at Mary.\\ Context 1: \textit{Mary and Tom forgot to invite John to the party.} vs. Context 2: \textit{The whole party is angry at Mary for forgetting the cake.}}

For example, for the above sentence, the \cs{} \textsc{Mary \& Tom} is more likely for Context 1, while BE\_ANGRY(e, AGENT=whole party, PATIENT=Mary, FOR=forgetting the cake) is more likely for Context 2. Both \cs{} interpretations are similarly likely with no additional context given.

\subsection{Entropy}
\begin{figure}[ht]
    \centering
    \includegraphics[width=0.45\textwidth]{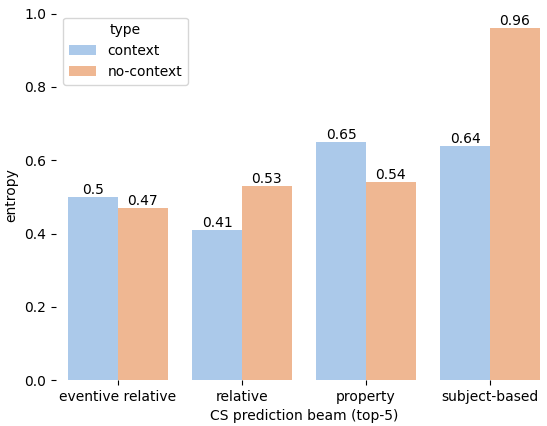}
    \caption{\dataset{} test set: Entropy over 4 different CS interpretation types, using top-5 beam predictions.
    }
    \label{fig:entropy_analysis}
\end{figure}

The entropy scores in Fig.~\ref{fig:entropy_analysis} show that context reduces the entropy over the semantic types of comparisons in the output beam of our model, for 2 categories, \textit{relative} and \textit{subject-based}. Interestingly, for \textit{eventive relative} and \textit{property}, entropy is slightly higher when shown the full context. 

The instances with the highest difference in entropy between the no-context and with-context setups, reveal some patterns. Cases where entropy is low for with-context but high for no-context tend to be extremely underspecified superlatives, usually in dialogue turns, where a sentence might simply say ``Which one is best?''. 
The \textit{eventive} and \textit{property} CS predictions have higher entropy for the full-context model, as it more frequently predicts eventive interpretations in its beam, compared to the no-context model.

\section{Related Work}

While superlatives have been widely studied in formal semantics, they have been largely neglected by NLP research, except for the following works. \citet{bos2006empirical} presented an automatic approach for predicting semantic interpretations of superlatives. 
For this purpose they annotated a corpus\footnote{We contacted the authors for access to the corpus, but unfortunately this corpus has been lost since the publication of their paper more than 15 years ago.} of attributive superlatives and their CS spans inside of a sentence.

\citet{scheible2008annotating} proposed an annotation scheme for identifying syntactic classes of superlatives and a semantic analysis of superlatives in terms of targets and CS. 
They further also automatically identified superlative surface forms and extracted targets and CS for a specific superlative sub-type \cite{scheible2009computational, scheible2012textwiki}. \citet{zhang2015semantic} also worked on identifying the targets and CS, using silver data from structured knowledge bases.

\citet{bakhshandeh2016learning} introduced a framework for comparative constructions, including superlatives, which is also able to model ellipsis, but it is limited to the sentence level. Similarly, \citet{pesahov2023qa} propose QA-based annotations for adjectives (which includes superlatives), but their annotation is constrained to a single sentence and does not include adverbial superlatives.

Multiple works have studied ambiguity, but none of them have specifically focused on superlatives: \citet{cui2022generalized} look at generalized quantifier ambiguity in multilingual NLI data, \citet{liu2023we} look at sentence ambiguity and its effect on entailment relations, and \citet{stengel2023zero} introduce a framework to translate ambiguous statements to formal representations. 

We extended upon previous superlative research by looking at a wider array of phenomena. Firstly, we are targeting \textbf{all} syntactic types of superlatives, 
while \citet{bos2006empirical} only analyzed the comparison sets of attributive 
superlatives and \citet{scheible2009computational} only analyzed predicative
superlatives. Most importantly, adverbial superlatives 
have not been studied in NLP. Lastly, we are extending the analysis of superlatives beyond the sentence boundary. While previous works did perform a (limited) analysis of implicit superlative phenomena inside of a sentence, we go beyond the sentence-based analysis and show how comparison sets are restricted by the broader domain of discourse. 

\section{Conclusion}
Superlative comparisons are interesting because their interpretation is closely tied to the context they appear in and because many of their components are often implicit. This paper provides a comprehensive study of superlatives, by proposing a new, unified annotation scheme and an annotated superlative dataset, \dataset. We further perform a set of experiments which analyze how models interpret superlative comparisons and how they are able to incorporate context restrictions. And with the help of the \dataset{} dataset we could investigate how ambiguity and context interact for superlative interpretations.

\bibliography{custom}

\begin{thebibliography}{39}
\providecommand{\natexlab}[1]{#1}

\bibitem[{Alshawi(1992)}]{alshawi1992core}
Hiyan Alshawi. 1992.
\newblock \emph{The core language engine}.
\newblock MIT press.

\bibitem[{Bakhshandeh et~al.(2016)Bakhshandeh, Wellwood, and Allen}]{bakhshandeh2016learning}
Omid Bakhshandeh, Alexis~Cornelia Wellwood, and James Allen. 2016.
\newblock Learning to jointly predict ellipsis and comparison structures.
\newblock In \emph{Proceedings of The 20th SIGNLL Conference on Computational Natural Language Learning}, pages 62--74.

\bibitem[{Bos and Nissim(2006)}]{bos2006empirical}
Johan Bos and Malvina Nissim. 2006.
\newblock An empirical approach to the interpretation of superlatives.
\newblock In \emph{Proceedings of the 2006 conference on empirical methods in natural language processing}, pages 9--17.

\bibitem[{Cui et~al.(2022)Cui, Hershcovich, and S{\o}gaard}]{cui2022generalized}
Ruixiang Cui, Daniel Hershcovich, and Anders S{\o}gaard. 2022.
\newblock Generalized quantifiers as a source of error in multilingual nlu benchmarks.
\newblock In \emph{Proceedings of the 2022 Conference of the North American Chapter of the Association for Computational Linguistics: Human Language Technologies}, pages 4875--4893.

\bibitem[{Elazar et~al.(2022)Elazar, Basmov*, Goldberg, and Tsarfaty}]{elazar2022text}
Yanai Elazar, Victoria Basmov*, Yoav Goldberg, and Reut Tsarfaty. 2022.
\newblock Text-based np enrichment.
\newblock \emph{Transactions of the Association for Computational Linguistics}, 10:764--784.

\bibitem[{Elazar and Goldberg(2019)}]{elazar2019s}
Yanai Elazar and Yoav Goldberg. 2019.
\newblock Where’s my head? definition, data set, and models for numeric fused-head identification and resolution.
\newblock \emph{Transactions of the Association for Computational Linguistics}, 7:519--535.

\bibitem[{Ettinger et~al.(2023)Ettinger, Hwang, Pyatkin, Bhagavatula, and Choi}]{ettinger2023you}
Allyson Ettinger, Jena Hwang, Valentina Pyatkin, Chandra Bhagavatula, and Yejin Choi. 2023.
\newblock “you are an expert linguistic annotator”: Limits of llms as analyzers of abstract meaning representation.
\newblock In \emph{Findings of the Association for Computational Linguistics: EMNLP 2023}, pages 8250--8263.

\bibitem[{Farkas and Kiss(2000)}]{farkas2000comparative}
Donka~F Farkas and Katalin~{\'E} Kiss. 2000.
\newblock On the comparative and absolute readings of superlatives.
\newblock \emph{Natural Language \& Linguistic Theory}, 18(3):417--455.

\bibitem[{Gawron(1995)}]{gawron1995comparatives}
Jean~Mark Gawron. 1995.
\newblock Comparatives, superlatives, and resolution.
\newblock \emph{Linguistics and Philosophy}, pages 333--380.

\bibitem[{Gerber and Chai(2010)}]{gerber2010beyond}
Matthew Gerber and Joyce Chai. 2010.
\newblock Beyond nombank: A study of implicit arguments for nominal predicates.
\newblock In \emph{Proceedings of the 48th Annual Meeting of the Association for Computational Linguistics}, pages 1583--1592.

\bibitem[{Geurts and van~der Sandt(1999)}]{geurts1999domain}
B~Geurts and RA~van~der Sandt. 1999.
\newblock Domain restriction.
\newblock \emph{Bosch, P.; Sandt, RA van der (ed.), Focus: Linguistic, Cognitive, and Computational Perspectives}, pages 268--292.

\bibitem[{Gundel et~al.(1993)Gundel, Hedberg, and Zacharski}]{gundel1993cognitive}
Jeanette~K Gundel, Nancy Hedberg, and Ron Zacharski. 1993.
\newblock Cognitive status and the form of referring expressions in discourse.
\newblock \emph{Language}, pages 274--307.

\bibitem[{Guti{\'e}rrez-Rexach(2006)}]{gutierrez2006superlative}
Javier Guti{\'e}rrez-Rexach. 2006.
\newblock Superlative quantifiers and the dynamics of context-dependence.
\newblock \emph{Where semantics meets pragmatics: The Michigan State University Papers}, pages 237--266.

\bibitem[{Heim(1999)}]{heim1999notes}
Irene Heim. 1999.
\newblock Notes on superlatives.
\newblock \emph{Ms., Massachusetts Institute of Technology}.

\bibitem[{Hirsch et~al.(2023)Hirsch, Pyatkin, Wolhandler, Caciularu, Shefer, and Dagan}]{hirsch2023revisiting}
Eran Hirsch, Valentina Pyatkin, Ruben Wolhandler, Avi Caciularu, Asi Shefer, and Ido Dagan. 2023.
\newblock Revisiting sentence union generation as a testbed for text consolidation.
\newblock In \emph{Findings of the Association for Computational Linguistics: ACL 2023}, pages 7038--7058.

\bibitem[{Hou et~al.(2013)Hou, Markert, and Strube}]{hou2013global}
Yufang Hou, Katja Markert, and Michael Strube. 2013.
\newblock Global inference for bridging anaphora resolution.
\newblock In \emph{Proceedings of the 2013 Conference of the North American Chapter of the Association for Computational Linguistics: Human Language Technologies}, pages 907--917.

\bibitem[{Huddleston and Pullum(2005)}]{huddleston2005cambridge}
Rodney Huddleston and Geoffrey Pullum. 2005.
\newblock The cambridge grammar of the english language.
\newblock \emph{Zeitschrift f{\"u}r Anglistik und Amerikanistik}, 53(2):193--194.

\bibitem[{Huddleston and Pullum(2002)}]{english.grammar.2002}
Rodney~D. Huddleston and Geoffrey~K. Pullum. 2002.
\newblock \emph{The Cambridge Grammar of the English Language}.
\newblock Cambridge University Press.

\bibitem[{Li et~al.(2017)Li, Su, Shen, Li, Cao, and Niu}]{li2017dailydialog}
Yanran Li, Hui Su, Xiaoyu Shen, Wenjie Li, Ziqiang Cao, and Shuzi Niu. 2017.
\newblock Dailydialog: A manually labelled multi-turn dialogue dataset.
\newblock In \emph{Proceedings of the Eighth International Joint Conference on Natural Language Processing (Volume 1: Long Papers)}, pages 986--995.

\bibitem[{Liu et~al.(2023)Liu, Wu, Michael, Suhr, West, Koller, Swayamdipta, Smith, and Choi}]{liu2023we}
Alisa Liu, Zhaofeng Wu, Julian Michael, Alane Suhr, Peter West, Alexander Koller, Swabha Swayamdipta, Noah~A Smith, and Yejin Choi. 2023.
\newblock We’re afraid language models aren’t modeling ambiguity.
\newblock In \emph{Proceedings of the 2023 Conference on Empirical Methods in Natural Language Processing}, pages 790--807.

\bibitem[{Mehta et~al.(2024)Mehta, Pyatkin, and Srikumar}]{mehta2024promptly}
Maitrey Mehta, Valentina Pyatkin, and Vivek Srikumar. 2024.
\newblock Promptly predicting structures: The return of inference.
\newblock \emph{arXiv preprint arXiv:2401.06877}.

\bibitem[{Misra(2022)}]{misra2022minicons}
Kanishka Misra. 2022.
\newblock minicons: Enabling flexible behavioral and representational analyses of transformer language models.
\newblock \emph{arXiv preprint arXiv:2203.13112}.

\bibitem[{Ni et~al.(2019)Ni, Li, and McAuley}]{ni2019justifying}
Jianmo Ni, Jiacheng Li, and Julian McAuley. 2019.
\newblock Justifying recommendations using distantly-labeled reviews and fine-grained aspects.
\newblock In \emph{Proceedings of the 2019 conference on empirical methods in natural language processing and the 9th international joint conference on natural language processing (EMNLP-IJCNLP)}, pages 188--197.

\bibitem[{Pesahov et~al.(2023)Pesahov, Klein, and Dagan}]{pesahov2023qa}
Leon Pesahov, Ayal Klein, and Ido Dagan. 2023.
\newblock Qa-adj: Adding adjectives to qa-based semantics.
\newblock In \emph{Proceedings of the Fourth International Workshop on Designing Meaning Representations}, pages 74--88.

\bibitem[{Price(1990)}]{price1990evaluation}
Patti Price. 1990.
\newblock Evaluation of spoken language systems: The atis domain.
\newblock In \emph{Speech and Natural Language: Proceedings of a Workshop Held at Hidden Valley, Pennsylvania, June 24-27, 1990}.

\bibitem[{Qi et~al.(2020)Qi, Zhang, Zhang, Bolton, and Manning}]{qi2020stanza}
Peng Qi, Yuhao Zhang, Yuhui Zhang, Jason Bolton, and Christopher~D Manning. 2020.
\newblock Stanza: A python natural language processing toolkit for many human languages.
\newblock In \emph{Proceedings of the 58th Annual Meeting of the Association for Computational Linguistics: System Demonstrations}, pages 101--108.

\bibitem[{Raffel et~al.(2020)Raffel, Shazeer, Roberts, Lee, Narang, Matena, Zhou, Li, and Liu}]{raffel2020exploring}
Colin Raffel, Noam Shazeer, Adam Roberts, Katherine Lee, Sharan Narang, Michael Matena, Yanqi Zhou, Wei Li, and Peter~J Liu. 2020.
\newblock Exploring the limits of transfer learning with a unified text-to-text transformer.
\newblock \emph{The Journal of Machine Learning Research}, 21(1):5485--5551.

\bibitem[{Roth and Frank(2015)}]{roth2015inducing}
Michael Roth and Anette Frank. 2015.
\newblock Inducing implicit arguments from comparable texts: A framework and its applications.
\newblock \emph{Computational Linguistics}, 41(4):625--664.

\bibitem[{Ruppenhofer et~al.(2010)Ruppenhofer, Sporleder, Morante, Baker, and Palmer}]{ruppenhofer2010semeval}
Josef Ruppenhofer, Caroline Sporleder, Roser Morante, Collin~F Baker, and Martha Palmer. 2010.
\newblock Semeval-2010 task 10: Linking events and their participants in discourse.
\newblock In \emph{Proceedings of the 5th International Workshop on Semantic Evaluation}, pages 45--50.

\bibitem[{Scheible(2008)}]{scheible2008annotating}
Silke Scheible. 2008.
\newblock Annotating superlatives.
\newblock In \emph{Proceedings of the Sixth International Conference on Language Resources and Evaluation (LREC'08)}.

\bibitem[{Scheible(2009)}]{scheible2009computational}
Silke Scheible. 2009.
\newblock \emph{Computational treatment of superlatives}.
\newblock Ph.D. thesis, The University of Edinburgh.

\bibitem[{Scheible(2010)}]{scheible2010smallest}
Silke Scheible. 2010.
\newblock The smallest, cheapest, and best: Superlatives in opinion mining.
\newblock page~52.

\bibitem[{Scheible(2012)}]{scheible2012textwiki}
Silke Scheible. 2012.
\newblock Textwiki: a superlative resource.
\newblock \emph{Language resources and evaluation}, 46:635--666.

\bibitem[{Schuler(2005)}]{schuler2005verbnet}
Karin~Kipper Schuler. 2005.
\newblock \emph{VerbNet: A broad-coverage, comprehensive verb lexicon}.
\newblock University of Pennsylvania.

\bibitem[{Stanley and Gendler~Szab{\'o}(2000)}]{stanley2000quantifier}
Jason Stanley and Zoltan Gendler~Szab{\'o}. 2000.
\newblock On quantifier domain restriction.
\newblock \emph{Mind \& Language}, 15(2-3):219--261.

\bibitem[{Stengel-Eskin et~al.(2023)Stengel-Eskin, Rawlins, and Van~Durme}]{stengel2023zero}
Elias Stengel-Eskin, Kyle Rawlins, and Benjamin Van~Durme. 2023.
\newblock Zero and few-shot semantic parsing with ambiguous inputs.
\newblock In \emph{The Twelfth International Conference on Learning Representations}.

\bibitem[{Szabolcsi(1986)}]{szabolcsi1986comparative}
Anna Szabolcsi. 1986.
\newblock Comparative superlatives.
\newblock In \emph{Papers in Theoretical Linguistics}, pages 245--266. MIT Working Papers in Linguistics.

\bibitem[{Zang et~al.(2020)Zang, Rastogi, Sunkara, Gupta, Zhang, and Chen}]{zang2020multiwoz}
Xiaoxue Zang, Abhinav Rastogi, Srinivas Sunkara, Raghav Gupta, Jianguo Zhang, and Jindong Chen. 2020.
\newblock Multiwoz 2.2: A dialogue dataset with additional annotation corrections and state tracking baselines.
\newblock In \emph{Proceedings of the 2nd Workshop on Natural Language Processing for Conversational AI}, pages 109--117.

\bibitem[{Zhang et~al.(2015)Zhang, Feng, Huang, Xu, Han, and Zhao}]{zhang2015semantic}
Sheng Zhang, Yansong Feng, Songfang Huang, Kun Xu, Zhe Han, and Dongyan Zhao. 2015.
\newblock Semantic interpretation of superlative expressions via structured knowledge bases.
\newblock In \emph{Proceedings of the 53rd Annual Meeting of the Association for Computational Linguistics and the 7th International Joint Conference on Natural Language Processing (Volume 2: Short Papers)}, pages 225--230.

\end{thebibliography}

\end{document}